\documentclass[letterpaper]{article} 
\usepackage[]{aaai24}  
\usepackage{times}  
\usepackage{helvet}  
\usepackage{courier}  
\usepackage[hyphens]{url}  
\usepackage{graphicx} 
\usepackage{natbib}  
\usepackage{caption} 
\usepackage{algorithm}
\usepackage{algorithmic}

\usepackage{newfloat}
\usepackage{listings}
\DeclareCaptionStyle{ruled}{labelfont=normalfont,labelsep=colon,strut=off} 
\floatstyle{ruled}
\newfloat{listing}{tb}{lst}{}
\floatname{listing}{Listing}
\usepackage{textcomp}
\usepackage{url}
\usepackage{color}
\usepackage{colortbl}
\usepackage{verbatim}
\usepackage{graphicx}

\usepackage{appendix} 

\usepackage{multicol}
\usepackage{multirow}
\usepackage{hhline}
\usepackage{makecell}
\usepackage{bm}
\usepackage{xcolor}

\usepackage{enumerate}
\usepackage{enumitem}
\usepackage{amsmath}

\usepackage{booktabs}
\usepackage{amssymb}

\usepackage{footnote}
\usepackage{caption}
\usepackage{etoolbox}

\newcommand{\ie}{\emph{i.e., }}
\newcommand{\eg}{\emph{e.g., }}

\title{Debiasing Multimodal Sarcasm Detection with Contrastive Learning}
\author{
   Mengzhao Jia\textsuperscript{\rm 1}, Can Xie\textsuperscript{\rm 1}, Liqiang Jing\textsuperscript{\rm 2}\thanks{Liqiang Jing is the corresponding author.}
}
\affiliations{
    \textsuperscript{\rm 1}Shandong University,
    \textsuperscript{\rm 2}University of Texas at Dallas\\

    \{jiamengzhao98, xiecan02, jingliqiang6\}@gmail.com
}

\begin{document}

\maketitle

\begin{abstract}
Despite commendable achievements made by existing work, prevailing multimodal sarcasm detection studies rely more on textual content over visual information. It unavoidably induces spurious correlations between textual words and labels, thereby significantly hindering the models' generalization capability. To address this problem, we define the task of out-of-distribution (OOD) multimodal sarcasm detection, which aims to evaluate models' generalizability when the word distribution is different in training and testing settings. 
Moreover, we propose a novel debiasing multimodal sarcasm detection framework with contrastive learning, which aims to mitigate the harmful effect of biased textual factors for robust OOD generalization. 
In particular, we first design counterfactual data augmentation to construct the positive samples with dissimilar word biases and negative samples with similar word biases. Subsequently, we devise an adapted debiasing contrastive learning mechanism to empower the model to learn robust task-relevant features and alleviate the adverse effect of biased words. 
Extensive experiments show the superiority of the proposed framework. 


\end{abstract}

\section{Introduction}
\begin{figure}
    \centering
    \includegraphics[width=\linewidth]{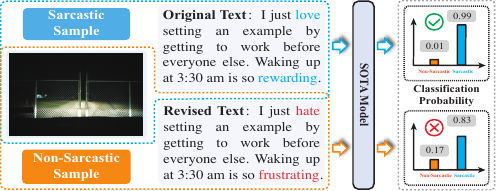}
    \caption{Prediction results of a sarcastic sample and the revised non-sarcastic sample by the SOTA model. The original sarcastic sample and rewritten non-sarcastic sample are included in blue and orange dashed boxes, respectively. 
    }
    \vspace{-1em}
    \label{fig:intro}
\end{figure}

With the rise of social media, individuals have increasingly embraced the use of ironic expressions in their posts on platforms such as Twitter\footnote{\url{https://twitter.com}.} and Weibo\footnote{\url{https://weibo.com}.}. Hence, the accurate detection of sarcastic/ironic expressions has become crucial for sentiment and opinion mining~\cite{DBLP:conf/acl/CaiCW19}. In pursuit of this aim, numerous researchers dedicate their efforts to identifying the underlying sarcastic/ironic semantics within social media posts.
Early studies~\cite{RiloffQSSGH13,PoriaCHV16} mainly concentrate on text-only approaches, which focus on recognizing sarcastic expressions in the textual content. While these endeavors have yielded impressive achievements, they are primarily centered around investigating sarcasm detection solely based on textual inputs. However, owing to the advancements in multimedia devices, individuals nowadays tend to express their emotions and opinions through multimodal social posts. Moreover, the visual content accompanying such posts often carries crucial cues for conveying sarcasm, as illustrated in Figure~\ref{fig:intro}.

Noticing this issue, nowadays the research interests have shifted to exploring the task of multimodal sarcasm detection (MSD), whose key objective is to accurately detect the inter- and intra-modal incongruities of someone’s implied sentiment expression within the given context. Early approaches incorporated fusion techniques that combined entire text and image features by concatenating operation~\cite{DBLP:conf/emnlp/PanL0Q020} or attention mechanism~\cite{DBLP:conf/iconip/GuptaSSSM21}. 
Despite their considerable progress, they overlook the possibility that sarcastic information may be expressed in some local segments of the text and certain regions of the image. Motivated by this, recent studies tried to employ advanced Graph Neural Networks (GNNs) to explore the local semantic relationships within the textual and visual modalities~\cite{DBLP:conf/acl/LiangLLY00PX22,DBLP:conf/mm/LiangLL00X21,DBLP:conf/aaai/QiaoJSCZN23,DBLP:conf/acl/JingSOJN23}. 

Despite the remarkable advancements made in MSD, existing studies still suffer from spurious correlations in the textual modality. Particularly, existing MSD models tend to place a heavier reliance on textual modality over visual modality~\cite{DBLP:conf/acl/LiangLLY00PX22,DBLP:conf/emnlp/PanL0Q020}. 
However, this over-reliance can be problematic as unreliable clues in the text (\eg biased words) can mislead the models, causing inaccurate predictions. 
By making slight word-level modifications to the original samples, the predicting accuracy of the MSD models exhibits substantial disparity. As shown in Figure~\ref{fig:intro}, even though the revised sample differs only marginally from the original sample in the text (love $\rightarrow$ hate, rewarding $\rightarrow$ frustrating), the two samples have opposite labels. Regrettably, the state-of-the-art~(SOTA) model~\cite{DBLP:conf/acl/LiangLLY00PX22} makes a wrong prediction for the revised sample, thereby highlighting the model's susceptibility to minor textual changes and revealing potential limitations in its ability to grasp robust task-related features (\ie the sarcastic semantics).
This outcome is attributed to the model's reliance on spurious correlations between biased words and the labels, rather than capturing the essential task-related features. 
Inevitably, deep neural networks are vulnerable to these spurious correlations due to the phenomenon of shortcut bias~\cite{DBLP:journals/natmi/GeirhosJMZBBW20}, hindering their generalization capability in out-of-distribution (OOD) scenarios, where the relationships between textual words and labels differ from those encountered during training. 
Consequently, we introduce a novel OOD MSD task to evaluate the models' true generalization ability by examining their capacity to effectively mitigate the adverse effects of unreliable biased factors in OOD scenarios.


In this light, we propose to mitigate the adverse impact caused by the biased words and encourage the learning of robust task-relevant features to enhance the model's generalization ability. 
To achieve this, we resort to contrastive learning~\cite{DBLP:conf/icml/ChenK0H20} that has demonstrated significant success in the representation learning field~\cite{DBLP:conf/aaai/YangACX22}.
In particular, we repel the feature representations of samples with similar biased words yet possess opposite labels (\ie negative samples). Meanwhile, we attract the feature representations of samples that contain dissimilar biased words but share the same label (\ie positive samples).
This allows the model to become more resistant to the biased factors induced by spurious correlations, while still preserving the meaningful associations between task-relevant semantic content and labels. 
The keys to achieving this objective are twofold: 1) acquiring negative samples with similar biased words, as well as positive samples comprising dissimilar biased words, and 2) learning robust task-relevant features. 


Aiming at this, we propose a novel debiasing MSD method with contrastive learning, DMSD-CL for short, which consists of the counterfactual data augmentation module and the adapted debiasing contrastive learning module. 
First, the counterfactual data augmentation module aims to construct both positive and negative samples. Leveraging in-context learning~\cite{DBLP:conf/nips/BrownMRSKDNSSAA20} and the ChatGPT~\footnote{\url{https://chat.openai.com}.} model, we tailor the augmentation manners to target at sarcastic and non-sarcastic samples separately.  
The adapted debiasing contrastive learning module, on the other hand, employs a re-weighted contrastive learning loss function to motivate the model to distinguish between negative samples that contain similar biased words, and to narrow the gap between dissimilar positive samples. 
Our contributions can be summarized as follows.
\begin{itemize}
    \item We define a novel OOD MSD task, which can assess the true generalization capability of the model based on an OOD testing scenario. 
    \item We propose a debiasing MSD framework with contrastive learning, in which counterfactual data augmentation is devised to construct positive and negative samples, and adapted debiasing contrastive learning is devised to make the model learn the robust feature. 
    \item We constructed an OOD testing set based on the existing MSD dataset. The experimental results on both the original and the OOD testing sets show the superiority of the proposed DMSD-CL. As a byproduct, we have released the source code and the constructed dataset\footnote{\url{https://sharecode.wixsite.com/dmsd-cl}.}. 
\end{itemize}


\section{Related Work}
\textbf{Sarcasm Detection}. 
Sarcasm detection has progressed from solely textual approaches to incorporating multimodal cues. With the proliferation of image-enabled platforms, researchers have increasingly sought to harness both visual and linguistic signals for discerning sarcasm. The seminal work by \citet{schifanella2016detecting} pioneered the MSD task by fusing textual and visual features. Lacking suitable datasets, \citet{DBLP:conf/acl/CaiCW19} compiled a new corpus from Twitter multimodal contents to facilitate data-driven methods. Subsequently, \citet{xu2020reasoning} and \citet{DBLP:conf/emnlp/PanL0Q020} emphasized modeling inter-modality incongruities, proposing decomposition networks and BERT-based architectures to capture contradictory cues. \citet{DBLP:conf/mm/LiangLL00X21} noted sarcastic intent often localizes in images and phrases, developing graph models relating localized text and visual concepts. However, these approaches still suffer from irrelevant visual regions. More recently, \citet{DBLP:conf/acl/LiangLLY00PX22} proposed constrained graph construction from detected objects and tokens to focus on pertinent local semantics. 
Nonetheless, the existing research overlooks the spurious correlation between textual modality and label. 

\noindent \textbf{Contrastive Learning}. 
Contrastive learning facilitates discriminative representation learning by identifying semantically similar instances among dissimilar ones. This technique has gained substantial attention across many domains, including computer vision~\cite{DBLP:conf/cvpr/He0WXG20} and information retrieval~\cite{DBLP:conf/iclr/XiongXLTLBAO21}. At its core, contrastive learning constructs positive and negative pairs via data augmentation for model optimization. For example, SimCLR~\cite{DBLP:conf/icml/ChenK0H20} generates positive and negative pairs using image augmentation techniques such as cropping, rotation, and Gaussian blurring.
More recently, the realm of contrastive learning has extended its reach to natural language tasks. However, unlike images, augmenting textual data necessitates an intricate consideration of syntactic and semantic structures. \citet{DBLP:conf/naacl/ZhangNWLZMNAX21} applied contrastive learning to text clustering, harnessing augmentations like back-translation and word replacement.
In departure from these endeavors, we apply contrastive learning to the domain of debiasing representation learning and introduces an innovative counterfactual data augmentation methodology for debiasing multimodal sarcasm detection.
\section{Method}
\subsection{Task Formulation}
\subsubsection{Standard MSD Task.} 
Given a training set that consists of $N$ samples $\mathcal{X}=\{x^1, x^2, \cdots, x^N\}$, where the $i$-th sample $x^i=\{T^i,v^i,y^i\}$ contains three elements. Therein, $T^i = \{t^i_1, t^i_2, ..., t^i_{n_i}\}$ refers to a sequence of $n_i$ textual tokens in the $i$-th sample and $v^i$ denotes the corresponding image. $y^i \in \{0, 1\}$ is the ground-truth label of sample $x^i$, where $y^i = 1$ indicates the sample is sarcastic and vice versa. 
Formally, given a sample $x^i$, we aim to design a model $\mathcal{F}$ that leverages both textual and visual information to determine whether $x^i$ implies any sarcasm,
\begin{equation}
    \hat{y}^i = \mathcal{F}(T^i,v^i |\Theta ),
\end{equation}
where $\Theta$ is the learnable model parameters and $\hat{y}^i$ refers to the prediction result.
We temporally omit the superscript $i$ that indexes the training samples for simplicity.

\subsubsection{OOD MSD Task.} 
To examine the true generalization capability of the model, we put forward the OOD MSD task to gauge the influence of spurious correlations on the MSD model. 
In particular, we built an OOD testing set by manually modifying samples from the biased MSD dataset, so that the word-label associations are markedly distinct compared to the training set. 
The difference in distributions between the training set and the OOD testing set enables measuring the effectiveness of the MSD model in mitigating the impact of spurious correlations. 

\subsection{MSD Model Initialization}
To implement the MSD model $\mathcal{F}$, we devise the commonly used model structure, which comprises the modal-specific encoding and multimodal fusion modules. 
\subsubsection{Modal-specific Encoding.}
We first acquire multimodal features from the samples by encoding the original visual and textual input via the following modal-specific encoders.

\textit{Textual Encoding.}
To better model the semantic information in the textual sentence, we feed it into the pre-trained language encoder RoBERTa~\cite{DBLP:journals/corr/abs-1907-11692}, which has gained appreciative results in multimodal language understanding tasks~\cite{DBLP:conf/emnlp/CaoLC022,DBLP:conf/iconip/GuptaSSSM21},  
\begin{equation}
    \mathbf{H}_t = \operatorname{RoBERTa}(T),
\end{equation}
where $\mathbf{H}_t \in \mathbb{R}^{n \times d_t}$ is the encoded textual representation. $n$ and $d_t$ represent the number of tokens in $T$ and the dimension of the hidden representations, respectively.

\textit{Visual Encoding.} 
In the MSD task, it is indispensable to understand the semantic content in the associated image as it possibly contains emotional incongruities with the text, which is a vital cue to reflect sarcasm. Aiming at this, we propose to exploit the visual information by a well-known pre-trained image encoder named ViT~\cite{DBLP:conf/iclr/DosovitskiyB0WZ21} as follows,
\begin{equation}
\mathbf{H}_v = \operatorname{ViT}(v),
\end{equation}
where the visual representation is denoted as $\mathbf{H}_v \in \mathbb{R}^{n_v \times d_v}$. Therein, $n_v$ and $d_v$ are the number of non-overlapping image patches and the hidden dimension of the ViT encoder, respectively.

\subsubsection{Multimodal Fusion}
After obtaining the visual and textual representations, we adopt the widely-used Cross-Attention~\cite{DBLP:conf/nips/VaswaniSPUJGKP17} mechanism to interact the information between the two modalities, so as to identify inter-modal incongruity.
Particularly, we treat the visual representation $\mathbf{H}_v$ as the \textit{query}, the textual representation $\mathbf{H}_t$ as both the \textit{key} and \textit{value} to form the cross attention. This guides the model to pay attention to textual features that are incongruous with image features. Formally,
\begin{equation}
    \mathbf{H}_m =\operatorname{Cross-Att}(\mathbf{H}_v,\mathbf{H}_t),
\end{equation}
where the interacted representation is denoted as $\mathbf{H}_m \in \mathbb{R}^{n_v \times d_t}$. We treat the first column of $\mathbf{H}_m$ as the multimodal representation vector $\mathbf{h} \in \mathbb{R}^{d_t}$ of $x$. Subsequently, we use projection and Softmax operations to obtain predicted sarcastic probability as follows,
\begin{equation}
    \hat{y} =  \sigma(\operatorname{MLP}(\mathbf{h})),
\end{equation}
where $\operatorname{MLP}$ is a Multi Layer Perceptron. The Softmax operator is denoted as $\sigma$. $\hat{y}$ represents the predicted result.

\subsection{DMSD-CL} 
In MSD, we expect the model to focus on the true inter- and intra- modal incongruity between two modalities to determine whether a sample contains sarcastic semantics. 
However, the presence of bias in the training samples gives rise to spurious correlations between samples and labels. For instance, certain words in the text may appear more frequently in sarcastic samples, despite lacking any inherent sarcastic sentiment.
These spurious correlations hinder the representation learning ability of existing models, introducing undesired bias factors into the learned representations, which contribute little to understanding the true sarcastic semantics. Consequently, the model's generalization ability to the OOD testing set is compromised.

To address this issue, we present DMSD-CL, a novel method comprising two modules: the counterfactual data augmentation module and the adaptive debiasing contrastive learning module. These modules effectively regulate the representation learning process, thereby encouraging the model to concentrate on genuine sarcasm-related semantic information.  

\begin{figure}
    \centering
    \includegraphics[width=\linewidth]{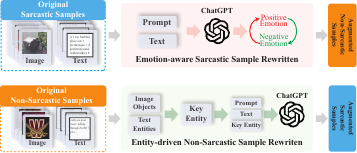}
    \caption{Counterfactual data augmentation for sarcastic and non-sarcastic samples. }
    \label{fig:aug}
    \vspace{-1em}
\end{figure}

\subsubsection{Counterfactual Data Augmentation.} 
In this section, we introduce a counterfactual data augmentation technique. 
It focuses on generating two types of counterfactual samples. The first type shares similar word biases with their corresponding factual (\ie original) counterparts while possessing opposite labels. Conversely, the second type shares different word biases compared to their corresponding counterparts while retaining identical labels. 
In particular, for the former, we design distinct data augmentation strategies, namely, emotion-aware sarcastic sample rewritten and entity-driven non-sarcastic sample rewritten, as shown in Figure~\ref{fig:aug}, tailored to the specific data characteristics of sarcastic and non-sarcastic samples, respectively. For the latter, we adopt a semantic-invariant data augmentation strategy.


\textit{Emotion-aware Sarcastic Sample Rewritten.} 
The sarcastic samples convey sarcastic semantics through intra- and inter-modal emotional incongruities~\cite{DBLP:conf/acl/CaiCW19,DBLP:conf/mm/LiangLL00X21}. Consequently, the textual modality of sarcastic samples usually exhibits emotions clearly and strongly. Leveraging this characteristic, we discover that flipping the sentiment polarity in the sarcastic sample can effectively neutralize its ironic semantics with minimal modifications. For instance, consider a sarcastic sample with a text statement: "What a fantastic hamburger!" while the accompanying image displays an unpleasant hamburger. In this case, the textual content might seem positive, but when combined with the image information, it reveals the true underlying sentiment of this sample is negative. By inverting the sentiment polarity, the statement transforms into "What a terrible hamburger!". The modification of a single word successfully eliminates the sarcastic element.

Accordingly, we invert the textual sentiment polarity of all sarcastic training samples to derive their counterfactual non-sarcastic counterparts. 
In light of the advanced understanding and sophisticated language generation capabilities of ChatGPT~\cite{DBLP:journals/corr/abs-2311-01477}, we employ it to reverse the sentiment polarity under minimal modifications to the original sentence. 
To be specific, we inform the ChatGPT model that the given text is ironic and ask the model to reversely change the sentiment in the text by making minor word-level modifications, so that the rewritten sample is no longer sarcastic. 
Similar to the previous work~\cite{DBLP:conf/acl/WuWY0FXQ23}, to improve the model's understanding of this instruction, we adopt the in-context learning scheme~\cite{DBLP:conf/nips/BrownMRSKDNSSAA20} and manually rewrite $K$ sarcastic samples' textual content as exemplars. The prompt is detailed in the Appendix. Ultimately, we obtain the augmented counterfactual non-sarcastic sample as follows,
\begin{equation}
    \Bar{T}_s \leftarrow \operatorname{ChatGPT}(Emo(T_s)),\\
\end{equation}
where $\Bar{T}_s$ is the rewritten text via ChatGPT using the prompt $Emo(T_s)$. $T_s$ denotes the text of factual sarcastic samples in the training set. 

\begin{table*}[t]
\centering
  \resizebox{\textwidth}{!}{
  \begin{tabular}{llccccccc}
    \hline
    \multicolumn{1}{c}{\multirow{2}*{MODALITY}} & \multicolumn{1}{c}{\multirow{2}*{METHOD}} &\multicolumn{1}{c}{\multirow{2}*{Acc(\%)}} & \multicolumn{3}{c}{\textit{F1-score}} & \multicolumn{3}{c}{\textit{\textit{Macro-average}}} \\
        \cline{4-9}
    ~ & & & Pre(\%) & Rec (\%) & F1 (\%) & Pre(\%) & Rec (\%) & F1 (\%) \\
\hline
\multirow{2}*{\textit{image-only}} 
~& ResNet~\cite{DBLP:conf/cvpr/HeZRS16} & 64.76 & 54.41 & 70.80 & 61.53 & 60.12 & 73.08 & 65.97\\

~& ViT~\cite{DBLP:conf/iclr/DosovitskiyB0WZ21} & 73.14 & 64.68 & 71.63 & 67.98 & 72.24 & 72.88 & 72.42\\ 
\hline
\multirow{4}*{\textit{text-only}} 
~&TextCNN~\cite{DBLP:conf/emnlp/Kim14}\textsuperscript{\textdagger} & 80.03 & 74.29 & 76.39 & 75.32 & 78.03 & 78.28 & 78.15\\
~& Bi-LSTM\textsuperscript{\textdagger}  & 81.90 &  76.66 & 78.42 & 77.53 & 80.97 & 80.13 & 80.55 \\
~& BERT~\cite{DBLP:conf/naacl/DevlinCLT19} & 83.85 & 78.72 & 82.27 & 80.22 & 81.31 & 80.87 & 81.09\\
~& RoBERTa~\cite{DBLP:journals/corr/abs-1907-11692} & 85.51 & 78.24 & 88.11 & 82.88 & 84.83 & 85.95 & 85.16 \\

\hline
\multirow{6}*{\textit{multi-modality}} 
~& HFM~\cite{DBLP:conf/acl/CaiCW19}\textsuperscript{\textdagger} & 83.44 & 76.57 & 84.15 & 80.18 & 79.40 & 82.45 & 80.90\\
~& Res-BERT~\cite{DBLP:conf/emnlp/PanL0Q020} & 84.30 & 79.02 & 82.48 & 80.71 & 83.54 & 83.99 & 83.74 \\
~& Att-BERT~\cite{DBLP:conf/emnlp/PanL0Q020}\textsuperscript{\textdagger} & 86.05 & 78.63 & 83.31 & 80.90 & 80.87 & 85.08 & 82.92\\
~& InCrossMGs~\cite{DBLP:conf/mm/LiangLL00X21}\textsuperscript{\textdagger} & 86.10 & 81.38 & 84.36 & 82.84 & 85.39 & 85.80 & 85.60\\
~& HCMKE~\cite{DBLP:conf/emnlp/LiuWL22} & 87.38 & 82.58 & 86.54 & 84.52 & 86.69 & 87.23 & 86.93\\
~& CMGCN~\cite{DBLP:conf/acl/LiangLLY00PX22}\textsuperscript{\textdagger} & 87.55 & 83.63 & 84.69 & 84.16 & 87.02 & 86.97 & 87.00\\
\rowcolor{gray!30}
~& DMSD-CL & \textbf{88.95} & \textbf{84.89} & \textbf{87.90} & \textbf{86.37} & \textbf{88.35} & \textbf{88.77} & \textbf{88.54}\\

\hline
\end{tabular}
}
\caption{
IID testing performance comparison among different methods on the MSD dataset. 
† indicates the results are cited from~\cite{DBLP:conf/acl/LiangLLY00PX22}, others are run by the open source codes. The best results are highlighted in boldface.
}

\label{table-iid}
\end{table*}

\textit{Entity-driven Non-Sarcastic Sample Rewritten.} 
In contrast to sarcastic samples, non-sarcastic samples often lack explicit and intuitive emotional expressions, making it challenging to directly revert the emotional polarity in these cases to get samples with opposite labels. Consequently, we adopt a distinct approach to obtain counterfactual samples for non-sarcastic instances. 
Through the analysis of existing sarcastic samples in the training set, we discern a common pattern among them, which is the presence of at least one explicit target entity to be satirized. Moreover, both the visual and textual modalities of the sarcastic sample revolve around this entity. Consequently, we first select the satirized target entity for counterfactual non-sarcastic sample rewritten.
In order to maintain the semblance of word distributions between the counterfactual and original samples, we propose to select the target entity from the pre-existing entities found in both the original visual and textual modalities. 
To achieve this, we begin by passing the text $T^n$ of non-sarcastic samples through a syntax parser\footnote{\url{https://www.nltk.org/}.} to extract all nouns, which serve as the textual entities $E_t$. Simultaneously, we leverage an advanced object detection technique, Faster-RCNN~\cite{DBLP:conf/nips/RenHGS15}, to identify a set of objects present in the image, constituting the visual entities $E_v$. 

Subsequently, we randomly pick out one entity in the overlapping entities of $E_t$ and $E_v$, and consider it as one of the target entities that the original sample focuses on. We can represent this process as follows, 
\begin{equation}
    e  \leftarrow \operatorname{Random}(E_t \cap E_v),
\end{equation}
where $e$ is the target entity derived by the random selection process $\operatorname{Random(\cdot)}$. 
Thereafter, we utilize ChatGPT to generate a text that satirizes the target entity $e$ by constructing a prompt based on $e$ as follows, 
\begin{equation}
     \Bar{T}_n \leftarrow \operatorname{ChatGPT}(Ent (e, T_n)),
\end{equation}
where the augmented text $\Bar{T}_n$ is obtained via the prompt $Ent(e, T_n)$ and ChatGPT. $Ent (e, T_n)$, constructed from the original non-sarcastic text $T_n$ and the target entity $e$, is derived to prompt ChatGPT to generate $\Bar{T}_n$, which has a sarcastic semantic under a similar word distribution with the original non-sarcastic text.
Similar to the previous section, we also manually rewrite $K$ textual content of non-sarcastic samples as demonstrations and prompt ChatGPT to rewrite other input data according to these examples. The concrete prompt can be found in the Appendix. 
By implementing the counterfactual non-sarcastic sample rewritten, we achieve to add sarcastic semantics to its textual content. During this process, we preserve the word distribution of the augmented text to be similar to the original text, which is achieved by constraining the target entity. 

\textit{Semantic-invariant Data Augmentation}. 
To get counterfactual samples sharing dissimilar biased words and identical labels with factual samples, we randomly choose some of four word-level modifications and perform it to the textual content of factual samples, motivated by EDA~\cite{EDA}. 
In particular, the four operations are: a) randomly replace a few words in the text with their synonyms, b) randomly insert the synonyms of random words into the text, c) randomly swap the positions of two words in the text, and d) randomly delete a few words from the text. 

Ultimately, for each sample $x$, we construct two counterfactual augmented samples, $\Bar{x}$ and $\widetilde{x}$, which have the opposite or the same sarcastic labels to $x$ as follows,
\begin{equation}
\begin{aligned}
    &\left\{\begin{array}{l}
    x = \{T,v,y\},  \\ 
    \Bar{x} = \{ \Bar{T},v,\Bar{y}\}, \\
    \widetilde{x} = \{\widetilde{T},v,y\},
   \end{array}\right.
\end{aligned}
\end{equation}
where $\Bar{y}$ represents the opposite label of $y$. 

\subsection{Adaptive Debiasing Contrastive Learning}
Based on the constructed triplet $(x,\Bar{x}, \widetilde{x})$, we introduce a contrastive learning framework to guide the model to perceive the sarcastic discrepancy between two samples of different labels but with similar word biases(\eg $x$ and $\Bar{x}$), as well as the sarcastic consistency between two samples of the same label but with dissimilar word biases(\eg $x$ and $\widetilde{x}$). 
Specifically, for each training sample $x$, we begin by defining its corresponding positive sample set $\mathcal{S}_p$ and negative sample set $\mathcal{S}_n$. 
Therein, $\mathcal{S}_p$ consists of samples (including factual and counterfactual samples) with the same label (except $x$ itself) in a mini-batch, and $\mathcal{S}_p$ consists of samples with the opposite label in a mini-batch. 
Based on the two sample sets, we aim to learn the robust latent representation of $x$ by attracting it towards the representations of samples in $\mathcal{S}_p$ and repels it from the representations of samples in $\mathcal{S}_n$ simultaneously as follows, 
\begin{equation}
   \mathcal{L}_c= \frac{1}{\left|\mathcal{S}_p\right|} \sum_{p \in \mathcal{S}_p}-\log \frac{\exp \left(\Phi (\mathbf{h}) \cdot \Phi (\mathbf{h}_p) / \tau\right)}{\sum_{n \in \mathcal{S}_n} \exp \left(\Phi (\mathbf{h}) \cdot \Phi (\mathbf{h}_n) / \tau\right)},\label{loss:cl}
\end{equation}
where $\tau$ is a temperature hyper-parameter. $\Phi$ denotes an MLP projection head which is widely used in contrastive learning. 
$\mathbf{h}_p$ and $\mathbf{h}_n$ is the final multimodal representation of the positive sample $p$ and negative sample $n$. 

However, Eqn.~\ref{loss:cl} weighs all positive and negative samples equally, which is sub-optimal in debiasing multimodal sarcasm detection. 
Intuitively, it poses greater difficulty for the model to distinguish the fundamentally different sarcastic semantics when samples with opposite labels exhibit greater biased word similarity. Conversely, when there is a significant disparity in biased word similarity for samples with the same labels, the model tends to make inconsistent predictions for them. 
Based on the aforementioned perceptions, we should encourage the model to focus more on 1) discriminating samples with similar word bias but opposite labels, and 2) narrowing the gap for samples with dissimilar word bias but the same label. 
To accomplish this, we employ the adaptive debiasing weighting strategy. Specifically, we first measure the biased similarity of two samples by the cosine similarity $r_{c} = cos<\mathbf{h}, \mathbf{h}_c> (c \in \{\mathcal{S}_p,\mathcal{S}_n\})$. 
$\mathbf{h}_c$ is the representation vector of sample $c$. 
$r_{c}$ denotes the cosine similarity between $\mathbf{h}$ and $\mathbf{h}_c$ which is regarded as the word bias similarity between $x$ and a sample $c$ from $\mathcal{S}_p$ or $\mathcal{S}_n$. Based on the similarity, the weight for each sample is as follows,
\begin{equation}
    w_{c} = \left\{\begin{array}{l}
    1- r_c, (c \in \mathcal{S}_p )\\
     1+r_{c}, (c \in \mathcal{S}_n ),
    \end{array}\right.
\end{equation}
where $w_c$ is the debiasing weight for sample $c$. Ultimately, the contrastive loss is refined as follows,
\begin{equation}
   \hat{\mathcal{L}_c}= \frac{1}{\left|\mathcal{S}_p\right|} \sum_{p \in \mathcal{S}_p}-\log \frac{w_p \cdot \exp \left(\Phi (\mathbf{h}) \cdot \Phi (\mathbf{h}_p) / \tau\right)}{\sum_{n \in \mathcal{S}_n} w_n \cdot  \exp \left(\Phi (\mathbf{h}) \cdot \Phi (\mathbf{h}_n) / \tau\right)},
\end{equation}
where $w_p$ and $w_n$ denote the weights assigned for a positive sample $p$ and a negative sample $n$, respectively. 

\subsection{Overall Optimization}
We employ the cross-entropy loss to optimize the sarcasm detection task as follows,
\begin{equation}
    \mathcal{L}_s = y \log (\hat{y}) + (1-y) \log (1-\hat{y}).
\end{equation}
Finally, we combine the classification and contrastive loss functions together to optimize the whole model as follows,
\begin{equation}
    \mathcal{L} = \mathcal{L}_s +  \lambda \hat{\mathcal{L}_c},
\end{equation}
where $\lambda$ is a hyper-parameter 
that controls the proportion of the two losses. 


\begin{table*}[t]
\centering
  \resizebox{\textwidth}{!}{
  \begin{tabular}{llccccccc}
    \hline
    \multicolumn{1}{c}{\multirow{2}*{MODALITY}} & \multicolumn{1}{c}{\multirow{2}*{METHOD}} &\multicolumn{1}{c}{\multirow{2}*{Acc(\%)}} & \multicolumn{3}{c}{\textit{F1-score}} & \multicolumn{3}{c}{\textit{\textit{Macro-average}}} \\
        \cline{4-9}
    ~ & & & Pre(\%) & Rec (\%) & F1 (\%) & Pre(\%) & Rec (\%) & F1 (\%) \\
\hline
\multirow{2}*{\textit{image-only}} 
~& ResNet~\cite{DBLP:conf/cvpr/HeZRS16} & 28.25 & 15.38 & 17.72 & 16.47 & 27.87 & 27.04 & 27.36\\

~& ViT~\cite{DBLP:conf/iclr/DosovitskiyB0WZ21} & 22.00 & 13.67 & 18.35 & 15.67 & 22.53 & 21.36 & 21.55\\ 
\hline
\multirow{4}*{\textit{text-only}} 
~&TextCNN~\cite{DBLP:conf/emnlp/Kim14}& 37.25 & 26.86 & 34.17 & 30.08 & 37.30 & 36.71 & 36.58\\
~& Bi-LSTM & 34.50 &  23.73 & 29.74 & 26.40 & 33.20 & 32.77 & 32.94 \\
~& BERT~\cite{DBLP:conf/naacl/DevlinCLT19} & 21.25 & 17.69 & 27.21 & 21.44 & 22.22 & 22.28 & 21.25\\
~& RoBERTa~\cite{DBLP:journals/corr/abs-1907-11692} & 29.50 & 15.16 & 17.08 & 16.07 & 28.07 & 27.34 & 27.64 \\

\hline
\multirow{6}*{\textit{multi-modality}} 
~& Res-BERT~\cite{DBLP:conf/emnlp/PanL0Q020} & 20.75 & 14.66 & 20.88 & 17.23 & 21.62 & 20.77 & 20.60 \\
~& Att-BERT~\cite{DBLP:conf/emnlp/PanL0Q020} & 28.25 & 21.58 & 31.01 & 25.45 & 27.50 & 26.46 & 26.69\\
~& HCMKE~\cite{DBLP:conf/emnlp/LiuWL22} & 37.50 & 27.88 & 36.70 & 31.69 & 37.90 & 37.36 & 37.04\\
~& CMGCN~\cite{DBLP:conf/acl/LiangLLY00PX22} & 34.25 & 27.27 & 39.87 & 32.39 & 35.52 & 35.22 & 34.20\\
\rowcolor{gray!30}
\rowcolor{gray!30}
~& DMSD-CL & \textbf{70.25} & \textbf{59.60} & \textbf{76.58} & \textbf{67.03} & \textbf{70.41} & \textbf{71.34} & \textbf{69.96}\\

\hline
\end{tabular}
}
\caption{\label{table:main-results} 
OOD testing performance comparison among different models on the MSD dataset. Since HFM and InCrossMGs did not publish the source codes, there are no corresponding results in the table. The best results are highlighted in boldface.}
\end{table*}

\begin{table}[th]
\centering
\begin{tabular}{l|cc|cc}
\hline
\multicolumn{1}{c}{\multirow{2}{*}{METHOD}} & \multicolumn{2}{|c|}{\textit{IID testing}} & \multicolumn{2}{c}{\textit{OOD testing}} \\
\cline{2-5}
~ & Acc & F1 & Acc & F1 \\
\hline
w/o-Image & 87.67  & 84.83 & 61.25 & 57.30 \\
w/o-Text  & 67.04 & 60.26 & 35.25 & 25.36 \\
w/o-Adapt & 88.33  & 85.71 & 64.75 & 62.59 \\
w/o-Contra & 88.37  & 85.47 & 25.75 & 15.38 \\
\hline
\rowcolor{gray!30}
\textbf{DMSD-CL} & \textbf{88.95} & \textbf{86.37}  & \textbf{70.25} & \textbf{67.03} \\
\hline
\end{tabular}
\caption{Experiment results of ablation study. }
\label{table:ablation-study}
\end{table}
\vspace{-1em}

\section{Experiments}
\subsection{Experimental Settings}
\subsubsection{Dataset.}
We conducted experiments on a publicly available multimodal sarcasm detection benchmark dataset \cite{DBLP:conf/acl/CaiCW19}.
The dataset comprises 24,635 samples, each comprising textual content, an associated image, and a corresponding label. Following the original setting, the sizes of the training set, development set, and testing set are 19,815, 2,410, and 2,409, respectively. 


\subsubsection{IID and OOD Settings.} 

As aforementioned, to test the genuine generalization ability of MSD models, we performed experiments on both Independent and Identically Distributed (IID) and OOD settings. In the IID setting, the data distributions of the training and testing set are similar. Conversely, in the OOD scenario, we created an OOD testing set in which the distribution of words over labels is different vastly from the training set. 
To this end, we employed proficient annotators to curate an OOD testing set, requiring that the samples should have similar textual content and opposite labels as original samples in the IID testing set.
Specifically, we instructed annotators to revise the textual content of the original sample with minimal changes to reverse its label. 
In addition, to enhance the revising process of the original sarcastic samples, we provided annotators with the explanation of why the sample is sarcastic, which is supplied by the prior work~\cite{DBLP:conf/aaai/Desai0A22}, to facilitate a deeper sarcasm semantics comprehension. 
Regarding the original non-sarcastic samples, we found that the samples with stronger emotional tendencies are more easily transformed into sarcastic variants. To systematically identify such candidates, we leveraged the NRC-VAD emotional dictionary\footnote{https://saifmohammad.com/WebPages/nrc-vad.html} to calculate samples' emotional intensities~\cite{DBLP:conf/emnlp/ZhongWM19} and select the original non-sarcastic samples with higher emotional intensities. Overall, we amassed a collection of the OOD testing set comprising 242 non-sarcastic and 158 sarcastic samples, each meticulously subjected to manual refinement and scrutiny. 

\subsubsection{Implementation Details.}
We adopted the pre-trained RoBERTa-base\footnote{https://huggingface.co/roberta-base}~\cite{DBLP:journals/corr/abs-1907-11692} model to generate embeddings for textual tokens. Besides, we employed the pre-trained ViT\footnote{https://github.com/lukemelas/PyTorch-Pretrained-ViT}~\cite{DBLP:conf/iclr/DosovitskiyB0WZ21} to produce embeddings for each visual region patch. We use $\operatorname{gpt3.5-turbo}$ for counterfactual data augmentation. The number of patches $n_v$ is set to $16$. The dimensions of the yielded textual and visual vectors $d_t$ and  $d_v$ are both $768$. Meanwhile, the hyperparameters $K$, $\tau$, and $\lambda$ are configured to $4$, $0.07$, and $0.9$, respectively. 
We used the Adam optimizer to optimize our model with a learning rate of $1e-5$. The mini-batch size is set to $16$ and the maximum number of epochs for training is set to $20$. 
Our evaluation employed Accuracy, Precision, Recall, Macro-average, and F1-score metrics to measure the model performance. 
\begin{figure*}
    \centering
    \includegraphics[width=0.9\textwidth]{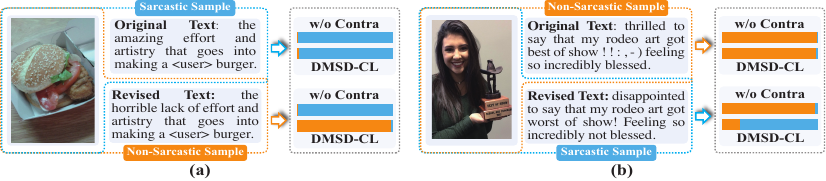}
    \vspace{-1em}
    \caption{Illustration of two samples in the IID testing set and OOD testing set. The bars on the right column of the image demonstrate the prediction distribution of models. The blue and orange colors denote probabilities of the sarcastic label and non-sarcastic label, respectively.}
    \label{fig:case}
    \vspace{-1.8em}
\end{figure*}

\subsection{Model Comparison}
We compared our model with a series of strong baselines, which can be broadly classified into three categories: 
1) \textit{Image-only Methods}. The sarcasm detection in these models relies solely on image input, incorporating methods such as \textbf{ResNet}~\cite{DBLP:conf/cvpr/HeZRS16} and \textbf{ViT}~\cite{DBLP:conf/iclr/DosovitskiyB0WZ21} to form the image-only sarcasm detection. 
2) \textit{Text-only Methods}. These methods purely rely on textual information for sarcasm detection, including \textbf{TextCNN}~\cite{DBLP:conf/emnlp/Kim14}, a deep learning model utilizing Convolutional Neural Networks for text classification; \textbf{Bi-LSTM}~\cite{DBLP:journals/neco/lstm}, a bidirectional Long Short-Term Memory network for text classification; \textbf{BERT}~\cite{DBLP:conf/naacl/DevlinCLT19} and \textbf{RoBERTa}~\cite{DBLP:journals/corr/abs-1907-11692}, which utilize the pre-trained BERT and RoBERTa models to detect sarcasm, respectively. 
3) \textit{Multimodal Methods}. These methods utilize both visual and textual information for sarcasm detection, including \textbf{HFM}~\cite{DBLP:conf/acl/CaiCW19}, which proposes a hierarchical model to fuse visual and textual features; \textbf{Res-BERT}~\cite{DBLP:conf/emnlp/PanL0Q020}, concatenating ResNet image features and BERT-based text features for sarcasm prediction; \textbf{Att-BERT}~\cite{DBLP:conf/emnlp/PanL0Q020}, exploring an inter-modality attention and a co-attention to model the incongruity of multimodal information; \textbf{InCrossMGs}~\cite{DBLP:conf/mm/LiangLL00X21}, a graph-based model to leverage the sarcastic relations from both intra- and inter-modality perspectives; \textbf{HCMKE}~\cite{DBLP:conf/emnlp/LiuWL22}, a novel hierarchical framework for sarcasm detection by exploring both the atomic-level congruity based on cross attention mechanism and the composition-level congruity based on graph neural network; and \textbf{CMGCN}~\cite{DBLP:conf/acl/LiangLLY00PX22}, a graph-based model exploring the sarcastic relations across objects of the image and tokens of the text.

We conducted experiments on both the IID and OOD testing sets. Table~\ref{table-iid} and Table \ref{table:main-results} show the comparison results of baseline models and DMSD-CL on IID and OOD scenarios, respectively. From the results, we can draw the following observations. 
1) DMSD-CL outperforms existing baselines in terms of all metrics on the IID testing set. This verifies the effectiveness of our proposed framework under the standard MSD task setting. 
2) DMSD-CL significantly surpasses all the baselines across all metrics on the OOD testing set. It shows the robustness and generalization ability of the proposed debiasing sarcasm detection framework,  which can overcome the data distribution (\ie biased words distribution) difference between the OOD testing set and the biased training set. 
3) The enhancement in performance achieved by DMSD-CL on the OOD testing set surpasses that observed on the IID testing set. The notable discrepancy in performance improvement can potentially be attributed to the debiasing capabilities of DMSD-CLis more easily observed on the OOD test set where the bias distribution difference between training and testing is larger.


\subsection{Ablation Study}
To verify the effectiveness of different components in our proposed framework, we compared it with the following variants. 
1) \textbf{w/o-Image} and 2) \textbf{w/o-Text}. To explore the effect of different modalities on sarcasm detection, we removed the visual and textual information from the framework, respectively. 
3) \textbf{w/o-Adapt}. To verify the importance of the devised adaptive debiasing weighting strategy, we excluded it in DMSD-CL by solely employing the standard contrastive learning method. 
4) \textbf{w/o-Contra}. To evaluate the role of adaptive debiasing contrastive learning with counterfactual data augmentation in sarcasm detection, we discarded the contrastive learning in this framework. 
 
Table \ref{table:ablation-study} summarizes the performance of DMSD-CL and its variants. From this table, we have the following observations.
1) DMSD-CL surpasses both w/o-Image and w/o-Text, demonstrating that removing either the visual or the textual information damages the sarcasm detection performance. 
2) w/o-Text performs better than w/o-Image, indicating textual content's stronger contribution to MSD than visual content, which also reveals the biased words in text may harm the model's generalization.
3) DMSD-CL consistently outperforms w/o-Adapt across different evaluation metrics. It verifies the effectiveness of the adaptive debiasing weighting strategy for debiasing sarcasm detection. 
4) w/o-Contra performs worse compared to DMSD-CL, possibly due because DMSD-CL can learn robust task-relevant representation while w/o-Contra cannot. This underscores the need for contrastive learning with counterfactual samples in debiasing sarcasm detection.

\subsection{Case Study}

To gain an intuitive comprehension of the DMSD-CL framework on the debiasing MSD task, we present experimental results of our DMSD-CL and its variant w/o-Contra. These results are illustrated through four distinct instances: two original samples from the IID testing set and two corresponding revised samples from the OOD testing set, as depicted in Figure~\ref{fig:case}. 
As can be seen, DMSD-CL yields accurate outcomes, whereas w/o-Contra falls short of achieving satisfying accuracy. 
Specifically, in case (a), w/o-Contra is misled by the biased words in the revised text, subsequently leading to an erroneous prediction. This can be attributed to that w/o-Contra grasps superficial associations in biased words and labels, without effectively capturing task-relevant attributes. Conversely, our DMSD-CL acquires sarcastic relevant representations through counterfactual data augmentation and adaptive debiasing contrastive learning. Consequently, DMSD-CL demonstrates robust generalization ability under the OOD scenario, revealing the necessity of mitigating the spurious correlations in the text and the label. 
A similar observation can be found in case (b). 

\section{Conclusion}
In this paper, we first analyze the spurious correlation between the text and label in MSD and then define a novel OOD MSD task to evaluate the generalization ability of models. Thereafter, to tackle this task, we propose a debiasing multimodal sarcasm detection framework with contrastive learning, which can discriminate the robust task-relevant feature from the superficial word bias. 
To measure the model performance in the OOD scenario, we constructed an OOD testing set by manual labeling. Extensive experiments on a public dataset demonstrate the superiority of the proposed framework on both IID and OOD settings.

\bibliography{aaai24}

\begin{thebibliography}{33}
\providecommand{\natexlab}[1]{#1}

\bibitem[{Brown et~al.(2020)Brown, Mann, Ryder, Subbiah, Kaplan, Dhariwal,
  Neelakantan, Shyam, Sastry, Askell, Agarwal, Herbert{-}Voss, Krueger,
  Henighan, Child, Ramesh, Ziegler, Wu, Winter, Hesse, Chen, Sigler, Litwin,
  Gray, Chess, Clark, Berner, McCandlish, Radford, Sutskever, and
  Amodei}]{DBLP:conf/nips/BrownMRSKDNSSAA20}
Brown, T.~B.; Mann, B.; Ryder, N.; Subbiah, M.; Kaplan, J.; Dhariwal, P.;
  Neelakantan, A.; Shyam, P.; Sastry, G.; Askell, A.; Agarwal, S.;
  Herbert{-}Voss, A.; Krueger, G.; Henighan, T.; Child, R.; Ramesh, A.;
  Ziegler, D.~M.; Wu, J.; Winter, C.; Hesse, C.; Chen, M.; Sigler, E.; Litwin,
  M.; Gray, S.; Chess, B.; Clark, J.; Berner, C.; McCandlish, S.; Radford, A.;
  Sutskever, I.; and Amodei, D. 2020.
\newblock Language Models are Few-Shot Learners.
\newblock In \emph{NeurIPS}.

\bibitem[{Cai, Cai, and Wan(2019)}]{DBLP:conf/acl/CaiCW19}
Cai, Y.; Cai, H.; and Wan, X. 2019.
\newblock Multi-Modal Sarcasm Detection in Twitter with Hierarchical Fusion
  Model.
\newblock In \emph{ACL}, 2506--2515. ACL.

\bibitem[{Cao et~al.(2022)Cao, Lee, Chong, and
  Jiang}]{DBLP:conf/emnlp/CaoLC022}
Cao, R.; Lee, R.~K.; Chong, W.; and Jiang, J. 2022.
\newblock Prompting for Multimodal Hateful Meme Classification.
\newblock In \emph{EMNLP}, 321--332. ACL.

\bibitem[{Chen et~al.(2020)Chen, Kornblith, Norouzi, and
  Hinton}]{DBLP:conf/icml/ChenK0H20}
Chen, T.; Kornblith, S.; Norouzi, M.; and Hinton, G.~E. 2020.
\newblock A Simple Framework for Contrastive Learning of Visual
  Representations.
\newblock In \emph{ICML}, 1597--1607. {PMLR}.

\bibitem[{Desai, Chakraborty, and Akhtar(2022)}]{DBLP:conf/aaai/Desai0A22}
Desai, P.; Chakraborty, T.; and Akhtar, M.~S. 2022.
\newblock Nice Perfume. How Long Did You Marinate in It? Multimodal Sarcasm
  Explanation.
\newblock In \emph{AAAI}, 10563--10571. {AAAI} Press.

\bibitem[{Devlin et~al.(2019)Devlin, Chang, Lee, and
  Toutanova}]{DBLP:conf/naacl/DevlinCLT19}
Devlin, J.; Chang, M.; Lee, K.; and Toutanova, K. 2019.
\newblock {BERT:} Pre-training of Deep Bidirectional Transformers for Language
  Understanding.
\newblock In \emph{NAACL-HLT}, 4171--4186. ACL.

\bibitem[{Dosovitskiy et~al.(2021)Dosovitskiy, Beyer, Kolesnikov, Weissenborn,
  Zhai, Unterthiner, Dehghani, Minderer, Heigold, Gelly, Uszkoreit, and
  Houlsby}]{DBLP:conf/iclr/DosovitskiyB0WZ21}
Dosovitskiy, A.; Beyer, L.; Kolesnikov, A.; Weissenborn, D.; Zhai, X.;
  Unterthiner, T.; Dehghani, M.; Minderer, M.; Heigold, G.; Gelly, S.;
  Uszkoreit, J.; and Houlsby, N. 2021.
\newblock An Image is Worth 16x16 Words: Transformers for Image Recognition at
  Scale.
\newblock In \emph{ICLR}. OpenReview.net.

\bibitem[{Geirhos et~al.(2020)Geirhos, Jacobsen, Michaelis, Zemel, Brendel,
  Bethge, and Wichmann}]{DBLP:journals/natmi/GeirhosJMZBBW20}
Geirhos, R.; Jacobsen, J.; Michaelis, C.; Zemel, R.~S.; Brendel, W.; Bethge,
  M.; and Wichmann, F.~A. 2020.
\newblock Shortcut learning in deep neural networks.
\newblock \emph{Nature Machine Intelligence}, 2(11): 665--673.

\bibitem[{Gupta et~al.(2021)Gupta, Shah, Shah, Syiemlieh, and
  Maurya}]{DBLP:conf/iconip/GuptaSSSM21}
Gupta, S.; Shah, A.; Shah, M.; Syiemlieh, L.; and Maurya, C. 2021.
\newblock FiLMing Multimodal Sarcasm Detection with Attention.
\newblock In \emph{{ICONIP}}, 178--186. Springer.

\bibitem[{He et~al.(2020)He, Fan, Wu, Xie, and
  Girshick}]{DBLP:conf/cvpr/He0WXG20}
He, K.; Fan, H.; Wu, Y.; Xie, S.; and Girshick, R.~B. 2020.
\newblock Momentum Contrast for Unsupervised Visual Representation Learning.
\newblock In \emph{CVPR}, 9726--9735. IEEE.

\bibitem[{He et~al.(2016)He, Zhang, Ren, and Sun}]{DBLP:conf/cvpr/HeZRS16}
He, K.; Zhang, X.; Ren, S.; and Sun, J. 2016.
\newblock Deep Residual Learning for Image Recognition.
\newblock In \emph{CVPR}, 770--778. {IEEE}.

\bibitem[{Hochreiter and Schmidhuber(1997)}]{DBLP:journals/neco/lstm}
Hochreiter, S.; and Schmidhuber, J. 1997.
\newblock Long Short-Term Memory.
\newblock \emph{Neural Comput.}, 9(8): 1735--1780.

\bibitem[{Jing et~al.(2023{\natexlab{a}})Jing, Li, Chen, Jia, and
  Du}]{DBLP:journals/corr/abs-2311-01477}
Jing, L.; Li, R.; Chen, Y.; Jia, M.; and Du, X. 2023{\natexlab{a}}.
\newblock {FAITHSCORE:} Evaluating Hallucinations in Large Vision-Language
  Models.
\newblock \emph{CoRR}, abs/2311.01477.

\bibitem[{Jing et~al.(2023{\natexlab{b}})Jing, Song, Ouyang, Jia, and
  Nie}]{DBLP:conf/acl/JingSOJN23}
Jing, L.; Song, X.; Ouyang, K.; Jia, M.; and Nie, L. 2023{\natexlab{b}}.
\newblock Multi-source Semantic Graph-based Multimodal Sarcasm Explanation
  Generation.
\newblock In \emph{{ACL}}, 11349--11361. ACL.

\bibitem[{Kim(2014)}]{DBLP:conf/emnlp/Kim14}
Kim, Y. 2014.
\newblock Convolutional Neural Networks for Sentence Classification.
\newblock In \emph{EMNLP}, 1746--1751. {ACL}.

\bibitem[{Liang et~al.(2021)Liang, Lou, Li, Gui, Yang, and
  Xu}]{DBLP:conf/mm/LiangLL00X21}
Liang, B.; Lou, C.; Li, X.; Gui, L.; Yang, M.; and Xu, R. 2021.
\newblock Multi-Modal Sarcasm Detection with Interactive In-Modal and
  Cross-Modal Graphs.
\newblock In \emph{Multimedia}, 4707--4715. {ACM}.

\bibitem[{Liang et~al.(2022)Liang, Lou, Li, Yang, Gui, He, Pei, and
  Xu}]{DBLP:conf/acl/LiangLLY00PX22}
Liang, B.; Lou, C.; Li, X.; Yang, M.; Gui, L.; He, Y.; Pei, W.; and Xu, R.
  2022.
\newblock Multi-Modal Sarcasm Detection via Cross-Modal Graph Convolutional
  Network.
\newblock In \emph{ACL}, 1767--1777. ACL.

\bibitem[{Liu, Wang, and Li(2022)}]{DBLP:conf/emnlp/LiuWL22}
Liu, H.; Wang, W.; and Li, H. 2022.
\newblock Towards Multi-Modal Sarcasm Detection via Hierarchical Congruity
  Modeling with Knowledge Enhancement.
\newblock In \emph{EMNLP}, 4995--5006. ACL.

\bibitem[{Liu et~al.(2019)Liu, Ott, Goyal, Du, Joshi, Chen, Levy, Lewis,
  Zettlemoyer, and Stoyanov}]{DBLP:journals/corr/abs-1907-11692}
Liu, Y.; Ott, M.; Goyal, N.; Du, J.; Joshi, M.; Chen, D.; Levy, O.; Lewis, M.;
  Zettlemoyer, L.; and Stoyanov, V. 2019.
\newblock RoBERTa: {A} Robustly Optimized {BERT} Pretraining Approach.
\newblock \emph{CoRR}, abs/1907.11692.

\bibitem[{Pan et~al.(2020)Pan, Lin, Fu, Qi, and
  Wang}]{DBLP:conf/emnlp/PanL0Q020}
Pan, H.; Lin, Z.; Fu, P.; Qi, Y.; and Wang, W. 2020.
\newblock Modeling Intra and Inter-modality Incongruity for Multi-Modal Sarcasm
  Detection.
\newblock In \emph{Findings of ACL}, 1383--1392. ACL.

\bibitem[{Poria et~al.(2016)Poria, Cambria, Hazarika, and Vij}]{PoriaCHV16}
Poria, S.; Cambria, E.; Hazarika, D.; and Vij, P. 2016.
\newblock A Deeper Look into Sarcastic Tweets Using Deep Convolutional Neural
  Networks.
\newblock In \emph{COLING}, 1601--1612.

\bibitem[{Qiao et~al.(2023)Qiao, Jing, Song, Chen, Zhu, and
  Nie}]{DBLP:conf/aaai/QiaoJSCZN23}
Qiao, Y.; Jing, L.; Song, X.; Chen, X.; Zhu, L.; and Nie, L. 2023.
\newblock Mutual-Enhanced Incongruity Learning Network for Multi-Modal Sarcasm
  Detection.
\newblock In \emph{{AAAI}}, 9507--9515. {AAAI} Press.

\bibitem[{Ren et~al.(2015)Ren, He, Girshick, and Sun}]{DBLP:conf/nips/RenHGS15}
Ren, S.; He, K.; Girshick, R.~B.; and Sun, J. 2015.
\newblock Faster {R-CNN:} Towards Real-Time Object Detection with Region
  Proposal Networks.
\newblock In \emph{NeurIPS}, 91--99.

\bibitem[{Riloff et~al.(2013)Riloff, Qadir, Surve, Silva, Gilbert, and
  Huang}]{RiloffQSSGH13}
Riloff, E.; Qadir, A.; Surve, P.; Silva, L.~D.; Gilbert, N.; and Huang, R.
  2013.
\newblock Sarcasm as Contrast between a Positive Sentiment and Negative
  Situation.
\newblock In \emph{EMNLP}, 704--714. {ACL}.

\bibitem[{Schifanella et~al.(2016)Schifanella, de~Juan, Tetreault, and
  Cao}]{schifanella2016detecting}
Schifanella, R.; de~Juan, P.; Tetreault, J.~R.; and Cao, L. 2016.
\newblock Detecting Sarcasm in Multimodal Social Platforms.
\newblock In \emph{Multimedia}, 1136--1145. {ACM}.

\bibitem[{Vaswani et~al.(2017)Vaswani, Shazeer, Parmar, Uszkoreit, Jones,
  Gomez, Kaiser, and Polosukhin}]{DBLP:conf/nips/VaswaniSPUJGKP17}
Vaswani, A.; Shazeer, N.; Parmar, N.; Uszkoreit, J.; Jones, L.; Gomez, A.~N.;
  Kaiser, L.; and Polosukhin, I. 2017.
\newblock Attention is All you Need.
\newblock In \emph{NeurIPS}, 5998--6008.

\bibitem[{Wei and Zou(2019)}]{EDA}
Wei, J.~W.; and Zou, K. 2019.
\newblock {EDA:} Easy Data Augmentation Techniques for Boosting Performance on
  Text Classification Tasks.
\newblock In \emph{EMNLP-IJCNLP}, 6381--6387. ACL.

\bibitem[{Wu et~al.(2023)Wu, Wang, Ye, Wu, Feng, Xu, and
  Qiao}]{DBLP:conf/acl/WuWY0FXQ23}
Wu, Z.; Wang, Y.; Ye, J.; Wu, Z.; Feng, J.; Xu, J.; and Qiao, Y. 2023.
\newblock OpenICL: An Open-Source Framework for In-context Learning.
\newblock In \emph{ACL}, 489--498. ACL.

\bibitem[{Xiong et~al.(2021)Xiong, Li, Tang, Liu, Bennett, Ahmed, and
  Overwijk}]{DBLP:conf/iclr/XiongXLTLBAO21}
Xiong, L. X.~C.; Li, Y.; Tang, K.; Liu, J.; Bennett, P.~N.; Ahmed, J.; and
  Overwijk, A. 2021.
\newblock Approximate Nearest Neighbor Negative Contrastive Learning for Dense
  Text Retrieval.
\newblock In \emph{ICLR}, 1--16. OpenReview.net.

\bibitem[{Xu, Zeng, and Mao(2020)}]{xu2020reasoning}
Xu, N.; Zeng, Z.; and Mao, W. 2020.
\newblock Reasoning with Multimodal Sarcastic Tweets via Modeling
  Cross-Modality Contrast and Semantic Association.
\newblock In \emph{ACL}, 3777--3786. ACL.

\bibitem[{Yang et~al.(2022)Yang, An, Cai, and Xu}]{DBLP:conf/aaai/YangACX22}
Yang, C.; An, Z.; Cai, L.; and Xu, Y. 2022.
\newblock Mutual Contrastive Learning for Visual Representation Learning.
\newblock In \emph{AAAI}, 3045--3053. {AAAI} Press.

\bibitem[{Zhang et~al.(2021)Zhang, Nan, Wei, Li, Zhu, McKeown, Nallapati,
  Arnold, and Xiang}]{DBLP:conf/naacl/ZhangNWLZMNAX21}
Zhang, D.; Nan, F.; Wei, X.; Li, S.; Zhu, H.; McKeown, K.~R.; Nallapati, R.;
  Arnold, A.~O.; and Xiang, B. 2021.
\newblock Supporting Clustering with Contrastive Learning.
\newblock In \emph{NAACL-HLT}, 5419--5430. ACL.

\bibitem[{Zhong, Wang, and Miao(2019)}]{DBLP:conf/emnlp/ZhongWM19}
Zhong, P.; Wang, D.; and Miao, C. 2019.
\newblock Knowledge-Enriched Transformer for Emotion Detection in Textual
  Conversations.
\newblock In \emph{EMNLP-IJCNLP}, 165--176. ACL.

\end{thebibliography}
\end{document}